\pdfoutput=1

\documentclass[11pt]{article}

\usepackage[]{naacl2021}

\usepackage{times}
\usepackage{latexsym}
\usepackage{multirow}
\usepackage{amsmath}
\usepackage{graphicx}
\usepackage{xspace}
\usepackage{xargs}
\usepackage{hhline}
\usepackage{comment}
\usepackage{booktabs}
\usepackage[linesnumbered,ruled]{algorithm2e}

\newcommand{\newsbias}{{\scshape NewsBias}}
\newcommand{\fakenews}{{\scshape FakeNews}}
\newcommand{\clickbait}{{\scshape Clickbait}}
\newcommand{\rumor}{{\scshape rumor}}
\newcommand{\propaganda}{{\scshape Propaganda}}
\newcommand{\buzzfeed}{{\scshape BuzzFeed}}
\newcommand{\politifact}{{\scshape PolitiFact}}
\newcommand{\roberta}{{RoBERTa}}
\newcommand{\unified}{\mbox{\scshape UnifiedM2}\xspace}
\newcommand{\ccnews}{{\scshape CC-News}}

\newcommand{\covidtwitter}{{\scshape CovidTwitter}}

\usepackage[T1]{fontenc}

\usepackage[utf8]{inputenc}

\usepackage{microtype}

\title{On Unifying Misinformation Detection}

\author{
Nayeon Lee$^\mathsection$\thanks{\hspace{2pt} Work partially done while interning at Facebook AI.}\quad Belinda Z. Li$^{\ddagger}$\thanks{\hspace{2pt} Work partially done while working at Facebook AI.}\quad Sinong Wang$^{\dagger}$ \\
\bf Pascale Fung$^\mathsection$\quad Hao Ma$^{\dagger}$\quad  Wen-tau Yih$^{\dagger}$\quad Madian Khabsa$^{\dagger}$ \\
$^\mathsection$Hong Kong University of Science and Technology \\ $^{\dagger}$Facebook AI \\ $^{\ddagger}$MIT CSAIL \\
\texttt {nyleeaa@connect.ust.hk},\quad
\texttt {mkhabsa@fb.com}
}

\begin{document}
\maketitle
\begin{abstract}
In this paper, we introduce \unified, a general-purpose misinformation model that jointly models multiple domains of misinformation with a single, unified setup. The model is trained to handle four tasks: detecting \textit{news bias}, \textit{clickbait}, \textit{fake news} and verifying \textit{rumors}. 
By grouping these tasks together, \unified learns a richer representation of misinformation, 
which leads to 
state-of-the-art or comparable performance across all tasks. Furthermore, we demonstrate that \unified's learned representation is helpful for few-shot learning of unseen misinformation tasks/datasets and model's generalizability to unseen events. 
\end{abstract}

\section{Introduction}


On any given day, $2.5$ quintillion bytes of information are created on the Internet, a figure that is only expected to increase in the coming years~\cite{marr_2019}.
The internet has allowed information to spread rapidly, and studies have found that misinformation spreads quicker and more broadly than true information~\cite{Vosoughi2018TheSO}.
It is thus paramount for misinformation detection approaches to be able to adapt to new, emerging problems in real time, without waiting for thousands of training examples to be collected. In other words, the \textit{generalizability} of such systems is essential.

\begin{figure}[t]
  \centering
  \includegraphics[width=\linewidth]{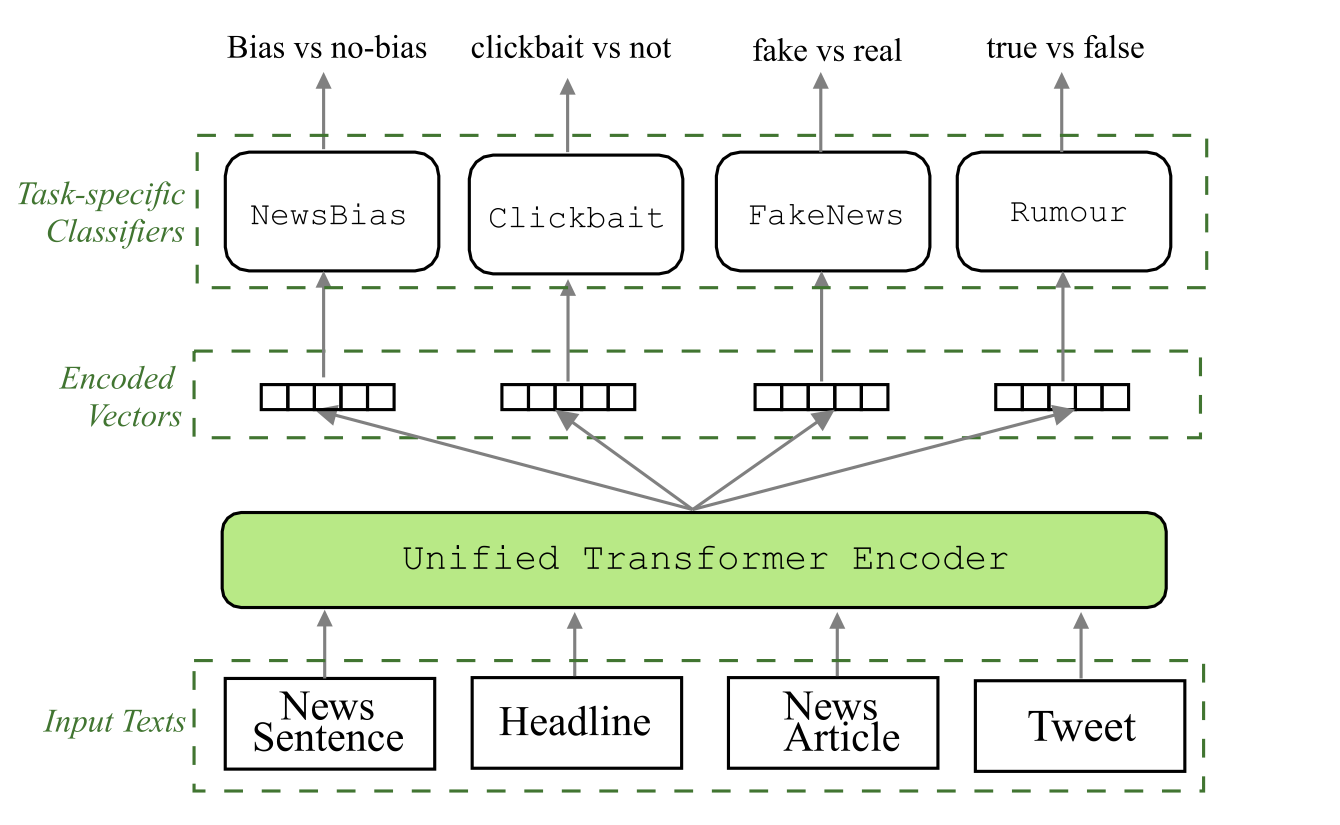}
  \caption{Architecture of our \unified~model}
  \label{fig:architecture}
\end{figure}

\begin{table*}[t]
\small
\centering
\begin{tabular}{lccccc}
\toprule
\textbf{Task} & 
\textbf{\begin{tabular}[c]{@{}c@{}}Dataset \\ Name\end{tabular}} &
\textbf{Granuarity} &
\textbf{\begin{tabular}[c]{@{}c@{}}Labels \\ (Positive/Negative)\end{tabular}} &

\textbf{Dataset Size} &
\textbf{\begin{tabular}[c]{@{}c@{}}Positive \\ Class Size\end{tabular}} \\
\midrule
\newsbias & BASIL & sentence & contains-bias/no-bias & 7,984 & 1,727 \\
\fakenews & Webis & article & fake/true & 1,627 & 363 \\
\rumor & PHEME & tweet & fake/true & 1,705 & 1,067 \\
\clickbait & Clickbait & headline & is-clickbait/not-clickbait & 19,538 & 4,761 \\ \bottomrule
\end{tabular}%
\caption{Summary of the four misinformation datasets we train on with~\unified.}
\label{table:seen_data_summaries}
\end{table*}

Misinformation detection is not well-studied from a generalizability standpoint. 
Misinformation can manifest in different forms and domains, i.e., fake news, clickbait, and false rumors, and
previous literature has mostly focused on building specialized models for a single domain~\cite{rubin2016fake,omidvar2018using,ma2018rumor}.
(Even prior literature on multi-tasking for misinformation~\cite{kochkina2018all} focuses more on using auxiliary tasks to boost performance on a \textit{single} task, rather than on all tasks.)
However, though these domains may differ in format (long articles vs. short headlines and tweets) and exact objective
(``is this fake'' vs.~``is this clickbait''), they have the same ultimate goal of deceiving their readers. As a result, their content often exhibits similar linguistic characteristics, such as using a sensational style to incite curiosity or strong emotional responses from readers. 
Furthermore, models trained on multiple tasks are more robust and less prone to overfitting to spurious domain-specific correlations.
Thus, unifying various domains of misinformation allows us to build a generalizable model that performs well across multiple domains/formats of misinformation. 

In this work, we propose Unified Misinfo Model (\unified), a misinformation detection model that uses multi-task learning~\cite{caruana1997multitask,maurer2016benefit,zhang2017survey}
to train on different domains of misinformation. 
Through a comprehensive series of empirical evaluations, we demonstrate that our approach is effective on \textit{all} tasks that we train on, improving $F_1$ in some cases by an absolute $\sim$8\%.
Moreover, we conduct ablation studies to more precisely characterize
how such positive transfer is attained. 
Beyond improvements on seen datasets,
we examine the generalizability of our proposed approach to unseen tasks/datasets and events.
This is highly applicable to real-world use cases, where obtaining new misinformation labels is costly and systems often wish to take down misinformation in real time.
Our experimental results indicate that our unified representation has better generalization ability over other baselines.  

\section{\unified}
\label{sec:model}
In this section, we describe the architecture and the training details for our proposed \unified~model.
\subsection{Architecture} Our proposed model architecture is a hard-parameter 
sharing multi-task learning model \cite{ruder2017overview}, 
where a single shared \roberta~\cite{liu2019roberta} encoder is used across all tasks.
\roberta~is a Transformer encoder pretrained with a masked-language-modeling objective on English Wikipedia and news articles (\ccnews), among other data. 
We additionally append task-specific multi-layer perceptron (MLP) classification heads following the shared encoder.
During multi-task training, the model sees examples from all datasets, and we jointly train the shared encoder with \textit{all} task-specific heads.
During inference time, we only use the classification head relevant to the inference-time task.
The overall architecture of the model is shown in Figure~\ref{fig:architecture}.

\subsection{Training} 
Our model training process consists of two steps. The first step is multi-task training of the shared \unified~encoder to learn a general misinformation representation. 
We jointly optimize for all tasks $t_1\cdots t_T$ by optimizing the sum of their task-specific losses $L_t$, where $L_t$ refers to the cross-entropy loss of the task-specific MLP classifiers. Our overall loss is defined as $L_{\mathrm{multi}} = \sum_{t = t_1\cdots t_T} L_{t}$. Note that since the dataset sizes are different, we over-sample from the smaller datasets to make the training examples roughly equal.
The second step is to fine-tune each task-specific heads again, similarly to the MT-DNN by \citet{liu2019multi}, to obtain the results reported in Table~\ref{table:all_result} and Table~\ref{table:few-shot}.

\section{Experiment}
Here, we provide experimental details (dataset, baselines, experimental setups) and results that empirically show the success of the proposed \unified~model.
\label{sec:experiments}
\begin{table*}[t]
\centering
\resizebox{0.65\linewidth}{!}{%
    \begin{tabular}{lcccccc}
    \toprule
    \multirow{2}[3]{*}{\textbf{Tasks}} & \multicolumn{2}{c}{\textbf{SoTA models}} & \multicolumn{2}{c}{\textbf{\roberta}} & \multicolumn{2}{c}{\textbf{\unified}} \\ 
    \cmidrule(lr){2-3} \cmidrule(lr){4-5} \cmidrule(lr){6-7} 
     & Acc & F1 & Acc & F1 & Acc & F1 \\ \midrule
    \newsbias~& N/A & 32.0\% $\vert$ 43.0\% & 72.8\% & 65.5\% & \textbf{81.0\%} & \textbf{70.2\%} \\
    \fakenews~& 58.0\% & 46.0\% & 84.3\% & \textbf{74.9\%} & \textbf{85.4\%} & 73.9\% \\
    \rumor~& 81.0\% & 80.0\% & 87.6\% & 86.9\% & \textbf{92.9\%} & \textbf{92.5\%} \\ 
    \clickbait~& 83.0\% & 57.0\% & 84.4\% & 77.4\% & \textbf{86.3\%} & \textbf{78.7\%} \\
    \bottomrule
    \end{tabular}
}
\caption{Results of single-task SoTA papers, the single-task \roberta~baseline, and our \unified~on all misinformation tasks. SoTA numbers for \newsbias, \fakenews~and \rumor~are from \citet{fan2019plain}, \citet{potthast2017stylometric}, and \citet{wu2019different}, respectively. \clickbait~numbers are from running the released code from \citet{omidvar2018using}. All the \roberta~and \unified~results are the averaged results of three seed runs.}
\label{table:all_result}
\end{table*}

\subsection{Misinformation Tasks/Dataset}
\label{sec:misinfo_tasks}
Table~\ref{table:seen_data_summaries} lists the 
four misinformation tasks/datasets we use to train \unified. They span various granularities and domains (articles, sentences, headlines and tweets) as well as various objectives (classifying veracity, bias and clickbaity-ness). 

\paragraph{\newsbias}
A task to classify whether a given sentence from a news article contains political bias or not. We adapt the BASIL~\cite{fan2019plain} dataset, which has bias-span annotations for lexical and informational bias within news articles. Using this dataset, we also include two auxiliary tasks related to political-bias detection: 1) \textit{bias type classification} -- given a biased sentence, the type of the bias (lexical vs informational) is classified; and 2) \textit{polarity detection} -- given a biased sentence, its polarity (positive, negative, neutral) is determined.

\paragraph{\fakenews}
An article-level fake news detection task that leverages the Webis~\cite{potthast2017stylometric} dataset annotated by professional journalists.

\paragraph{\rumor}
A task to verify the veracity of a rumor tweet. The PHEME dataset~\cite{zubiaga2016analysing}, which contains rumor tweets with their corresponding reply tweets (social engagement data), is used for this task. We only use the text of the source rumor tweet since we focus on learning a good representation for misinformation \textit{text}. Originally, there were three class labels (true, false, unverified); however, following other literature~\cite{derczynski2017semeval, wu2019different}, we report the binary version, excluding the unverified label.

\paragraph{\clickbait} A task to detect the clickbaity-ness of news headlines, which refers to sensational headlines that might deceive and mislead readers. For this task, we use the dataset from the Clickbait Challenge.\footnote{https://www.clickbait-challenge.org/. There are two versions of the labeled dataset, but we only use the larger one.}

\subsection{Baseline Models}
\paragraph{State-of-the-Art Models}
For each misinformation task, we report and compare our approach to the SoTA models from \citet{fan2019plain} for \newsbias,\footnote{They report bias-detection performance separately on the ``lexical-bias vs. no-bias'' setting and ``informational-bias vs. no-bias'' setting. In our experiments, we treat both lexical-bias and informational-bias to be ``contains-bias'' class, and conduct one unified experiment.} \citet{potthast2017stylometric} for \fakenews, \citet{wu2019different} for \rumor~and \citet{omidvar2018using} for \clickbait. 
    
\paragraph{\roberta-based Baselines} In addition to each task's published SoTA model, we create \roberta-based models by fine-tuning \roberta~to each individual task.

\subsection{Experimental Setup}
\paragraph{Training Details}
We ran all our experiments for 3 times with different shots, and report the average. Our \unified~ model is based on RoBERTa-large model which has 355M parameters. 

We used the Adam optimizer~\cite{kingma2014adam} with a mini-batch size of $32$. The learning rate was set to 5e-6 with linear learning rate decay. 
The maximum epoch count was 15, with early stopping patience set to 5. The maximum sequence length of input was set to $128$.
These parameters were obtained by performing grid-search over our validation loss. We search within the following hyper-parameter bounds: $LR=\{5e-5, 5e-6, 5e-7\}$, $batch=\{16,32\}$. 

\paragraph{Training Details for few-shot experiments}
We did not do any parameter searching for these few-shot experiments. We kept all the training details and parameters the same to the training details that are state above.

\paragraph{Computing Infrastructure}
We ran all experiments with 1 NVIDIA TESLA V100 GPU with 32 GB of memory.

\subsection{Main Results}
Table \ref{table:all_result} presents the results of our proposed unified model, \unified, along with the two groups of baseline models.
\unified achieves better or comparable results over both baselines for \textit{all four} misinformation tasks.
The improvement is especially prominent on the \newsbias~and \rumor~tasks, where we see an $~8\%$ and $~5\%$ improvement in accuracy, respectively. 

\begin{table}[]
\resizebox{\linewidth}{!}{%
\begin{tabular}{llcc}
\toprule
\textbf{\#} & \textbf{Task Combination} & \textbf{Acc} & \textbf{F1} \\ \midrule
1 & \rumor \ ST \roberta~& 87.6\% & 86.9\% \\ \hline
\multirow{3}{*}{2} & \rumor, \newsbias & 85.9\% & 85.4\% \\
 & \rumor, \clickbait & 88.8\% & 87.8\% \\
 & \rumor, \fakenews & 78.8\% & 78.7\% \\ \hline
\multirow{3}{*}{3} & \newsbias, \fakenews, \rumor & 88.2\% & 87.5\% \\
 & \newsbias, \rumor, \clickbait & 91.8\% & 90.5\% \\
 & \fakenews, \rumor, \clickbait & 88.8\% & 87.8\% \\ \hline
4 & \unified~& \textbf{92.9\%} & \textbf{92.5\%} \\
\bottomrule
\end{tabular} %
}
\caption{Ablation study for understanding which task(s), when trained in combination with ~\rumor, are most beneficial when evaluated on~\rumor.}
\label{table:ablation}
\end{table}

\begin{table*}[]
\resizebox{\linewidth}{!}{%
    \begin{tabular}{lccccccccc}
    \toprule
    \multirow{2}{*}{\textbf{Tasks}} & \multicolumn{3}{c}{\textbf{10 examples}} & \multicolumn{3}{c}{\textbf{25 examples}} & \multicolumn{3}{c}{\textbf{50 examples}} \\ 
    \cmidrule(lr){2-4} \cmidrule(lr){5-7} \cmidrule(lr){8-10} 
     & \textbf{Vanilla} & \textbf{ST average} & \textbf{\unified} & \textbf{Vanilla} & \textbf{ST average} & \textbf{\unified} & \textbf{Vanilla} & \textbf{ST average} & \textbf{\unified} \\ \midrule
    Propaganda & 45.10\% & 55.50\% & \textbf{56.19\%} & 56.4\% & 60.7\% & \textbf{62.5\%} & 56.4\% & 65.5\% & \textbf{72.9\%} \\
    Fake News Article & 32.97\% & 38.70\% & \textbf{42.42\%} & 35.0\% & \textbf{58.1\%} & 53.1\% & 35.0\% & 67.3\% & \textbf{74.2\%} \\
    Fake News Title & 34.13\% & \textbf{64.08\%} & 55.36\% & 33.9\% & 66.6\% & \textbf{67.0\%} & 33.9\% & \textbf{73.5\%} & 71.4\% \\
    Covid Check-worthy Twitter & 52.17\% & 55.22\% & \textbf{61.70\%} & 37.3\% & 62.5\% & \textbf{66.4\%} & 37.3\% & 65.6\% & \textbf{73.2\%} \\
    Covid False Twitter Claim & 46.93\% & 48.01\% & \textbf{54.25\%} & 46.4\% & 51.2\% & \textbf{56.3\%} & 46.4\% & 54.0\% & \textbf{59.7\%} \\ \midrule
    Average & 42.26\% & 52.30\% & \textbf{53.98\%} & 41.80\% & 59.84\% & \textbf{61.05\%} & 41.80\% & 65.19\% & \textbf{70.27\%} \\ \bottomrule
    \end{tabular}%
}
\caption{Macro-F1 scores of the few-shot experiment with 10, 25, and 50 examples on unseen misinformation-related datasets/tasks. The following datasets are used for each task: \propaganda: Propaganda detection, \politifact: fake news article detection, \buzzfeed: fake news title detection, \covidtwitter: Check-worthy twitter detection and false twitter claim detection (this dataset is used for two tasks) }
\label{table:few-shot}
\end{table*}

\subsection{Task Ablation Study}
We conduct an ablation study to better understand how other tasks help in our multitask framework.
One question we ask is what \textit{kinds} of tasks benefit the most from being trained together. Namely, how well do more ``similar'' vs. more ``different'' kinds of task transfer to each other?

Specifically, we use the~\rumor~dataset as a case study\footnote{Other datasets show similar findings.}. We train on multiple task combinations and evaluate their performance on \rumor. Results are shown in Table~\ref{table:ablation}.
Note that adding \fakenews~alone to single-task \roberta, or \newsbias, actually \textit{hurts} performance, indicating that multi-task learning is not simply a matter of data augmentation.
We hypothesize that the drop is due to~\fakenews~being the least similar in format and style to~\rumor. 
Qualitatively, we compare examples from~\fakenews~and \clickbait~(the most helpful dataset) to \rumor. 
Examples from~\fakenews~are long documents with a mix of formal and sensational styles, whereas 
\clickbait~contains short, sensational sentences.

However, 
as the model is trained on more datasets,
adding the less similar \fakenews~task actually \textit{improves} overall performance 
($90.5\to 92.5$ F1 in three datasets), despite hurting the model trained on \rumor~only ($86.9\to 78.7$ F1).
We hypothesize this is due, in part, to including more diverse sources of data, which improves the \textit{robustness} of the model to different types of misinformation.



\section{Generalizability Analysis}

New types, domains, and subjects of misinformation arise frequently. Promptly responding to these new sources is challenging, as they can spread widely 
before there is time to collect sufficient task-specific
training examples. 
For instance, the rapid spread of COVID-19 was accompanied by equally fast spread of large quantities of misinformation~ \cite{joszt2020ajmc,kouzy2020coronavirus}.

Therefore, we carry out experiments to evaluate the generalization ability of \unified~representation to unseen misinformation (i) tasks/datasets and (ii) events. The first experiment is about fast adaption ability (few-shot training) to handle a new task/dataset, whereas the second experiment is about the model's ability to perform well on events unseen during training. 

\subsection{Unseen Task/Dataset Generalizability}
\label{subsection:unseen_task}
\paragraph{Dataset}
We evaluate using the following four unseen datasets: \propaganda~\cite{da2019fine}, which contains 21,230 propaganda and non-propaganda sentences, with the propaganda sentences annotated by fine-grained propaganda technique labels, such as ``Name calling'' and  ``Appeal to fear''; \politifact~\cite{shu2019beyond}, which contains 91 true and 91 fake news \textit{articles} collected from PolitiFact's fact-checking platform; 
\buzzfeed~\cite{shu2019beyond}, which contains 120 true and 120 fake news \textit{headlines} collected from BuzzFeed's fact-checking platform;
and \covidtwitter~\cite{alam2020fighting}, which contains
504 COVID-19-related tweets.
For our experiment, we use two of the annotations: 1) \textit{Twitter Check-worthiness}: does the tweet contain a verifiable factual claim? 2) \textit{Twitter False Claim}: does the tweet contain false information?

\paragraph{Few-shot Experiments}
We compare the few-shot performance of \unified~against off-the-shelf \roberta~and single-task \roberta. 
For each unseen dataset, a new MLP classification head is trained on top of the \roberta~encoder, in a few-shot manner. Given $N_{d}$ to be the size of the given dataset $d$, we train the few-shot classifiers with $k$ randomly selected samples and evaluate on the remaining $N-k$ samples. We test with $k={10,25,50}$. Note that for single-task \roberta, we report the average performance across the four task-specific models (ST average).

As shown in Table~\ref{table:few-shot}, our \unified~encoder can quickly adapt to new tasks, even with very little in-domain data. 
While both the single-task models and \unified~significantly outperform vanilla \roberta, \unified~further outperforms the single-task models, indicating that multi-task learning can aid task generalizability.

\subsection{Unseen Event Generalizability}
\paragraph{Dataset}
We use the previously introduced \rumor~dataset, which includes nine separate events, for this experiment. A group of works~\cite{kochkina2018all,li2019rumor,yu2020coupled} have used this dataset in a leave-one-event-out cross-validation setup (eight events for training and one event for testing) to take event generalizability into consideration in their model evaluation. We conduct a supplementary experiment following this evaluation setup for the completeness of our analysis.  

\paragraph{Experiment}
First, we train the \unified~encoder \textit{without} \rumor~data, and then fine-tune and evaluate in the leave-one-event-out cross-validation setup. Note that we re-train the \unified~encoder to ensure that it has no knowledge of the left-out-event testset. Results in Table~\ref{table:rumor_event_test} show that our proposed method outperforms two recent SoTA models~\cite{li2019rumor,yu2020coupled} by an absolute $16.44\%$ and $25.14\%$ in accuracy. This indicates that unified misinformation representations are helpful in event generalizability as well. 

\begin{table}[]
\centering
\small
\begin{tabular}{ccc}
\toprule
\textbf{Model} & \textbf{Acc} & \textbf{F1} \\ \midrule
SoTA'19 \cite{li2019rumor} & 48.30\% & 41.80\% \\
SoTA'20 \cite{yu2020coupled} & 39.60\% & 46.60\% \\ \midrule
Vanilla & 47.07\% & 33.90\% \\
\unified~ & 64.74\% & 44.74\% \\ \bottomrule
\end{tabular}
\caption{Average acc and macro-F1 scores from leave-one-event-out cross-validation setup for \rumor~task. }
\label{table:rumor_event_test}
\end{table}

\section{Related Work}
\label{sec:related_works}

Existing misinformation works take three main approaches: \textit{Content-based} approaches examine the language of a document only. Prior works have looked at 
linguistic features such as hedging words and emotional words~\cite{rubin2016fake,potthast2017stylometric,rashkin2017truth,wang2017liar}. \textit{Fact-based} approaches leverage evidence from external sources (e.g., Wikipedia, Web) to determine the truthfulness of the information~\cite{Etzioni2008,wu2014toward,ciampaglia2015computational,popat2018declare,thorne2018fever,nie2019combining}. 
Finally, \textit{social-data-based} approaches use the surrounding social data--such as the credibility of the authors of the information \cite{long2017fake,kirilin2018exploiting,li2019rumor} or social engagement data~\cite{derczynski2017semeval,ma2018rumor,kwon2013prominent,volkova2017separating}.

Though prior works have explored multi-task learning within misinformation, they have focused exclusively on one domain.
These works try to predict two different labels on the same set of examples from a single~\cite{kochkina2018all} or two closely-related datasets~\cite{wu2019different}.
In contrast, our proposed approach crosses not just \textit{task} or \textit{dataset} boundaries, but also \textit{format} and \textit{domain} boundaries. 
Furthermore, prior works focus on using an auxiliary task to boost the performance of the main task, while we focus on using multitasking to \textit{generalize} across many domains.
Thus, the focus of this work is not the multitask paradigm, but rather the unification of the various domains, using multitasking.

\section{Conclusion}
\label{sec:conclusion} 

In this paper, we introduced~\unified, 
which unifies multiple domains of misinformation with a single multi-task learning setup. We empirically showed that such unification improves the model's performance against strong baselines, and achieves new state-of-the-art results. 
Furthermore, we show that \unified~can generalize to out-of-domain misinformation tasks and events, 
and thus can serve as a good starting point for others working on misinformation.

\bibliography{anthology,custom}
\bibliographystyle{acl_natbib}

\clearpage
\appendix

\end{document}